# Deep Representation Learning of Patient Data from Electronic Health Records (EHR): A Systematic Review

Yuqi Si[a], Jingcheng Du[a], Zhao Li[a], Xiaoqian Jiang[a], Timothy Miller[b], Fei Wang[c], W. Jim Zheng[a], Kirk Roberts[a]

[a] School of Biomedical Informatics, The University of Texas Health Science Center at Houston, Texas, USA
[b] Computational Health Informatics Program (CHIP), Boston Children's Hospital and Harvard Medical School, Massachusetts, USA
[c] Department of Population Health Sciences. Weill Cornell Medicine, Cornell University, New York, USA

Corresponding Author:
Kirk Roberts, PhD
School of Biomedical Informatics
University of Texas Health Science Center at Houston
7000 Fannin St. #600
Houston TX 77030
kirk.roberts@uth.tmc.edu

## ABSTRACT

### Objectives:

*Patient representation learning refers to learning a dense mathematical representation of a patient that encodes meaningful information from Electronic Health Records (EHRs). This is generally performed using advanced deep learning methods. This study presents a systematic review of this field and provides both qualitative and quantitative analyses from a methodological perspective.*

### Methods:

*We identified studies developing patient representations from EHRs with deep learning methods from MEDLINE, EMBASE, Scopus, the Association for Computing Machinery (ACM) Digital Library, and the Institute of Electrical and Electronics Engineers (IEEE) Xplore Digital Library. After screening 363 articles, 49 papers were included for a comprehensive data collection.*

### Results:

*Publications developing patient representations almost doubled each year from 2015 until 2019. We noticed a typical workflow starting with feeding raw data, applying deep learning models, and ending with clinical outcome predictions as evaluations of the learned representations. Specifically, learning representations from structured EHR data was dominant (37 out of 49 studies). Recurrent Neural Networks were widely applied as the deep learning architecture (Long short-term memory: 13 studies, Gated recurrent unit: 11 studies). Learning was mainly performed in a supervised manner (30 studies) optimized with cross-entropy loss. Disease prediction was the most common application and evaluation (31 studies). Benchmark datasets were mostly unavailable (28 studies) due to privacy concerns of EHR data, and code availability was assured in 20 studies.*

### Discussion & Conclusion:

*The existing predictive models mainly focus on the prediction of single diseases, rather than considering the complex mechanisms of patients from a holistic review. We show the importance and feasibility of learning comprehensive representations of patient EHR data through a systematic review. Advances in patient representation learning techniques will be essential for powering patient-level EHR analyses. Future work will still be devoted to leveraging the richness and potential of available EHR data. Reproducibility and transparency of reported results will hopefully improve. Knowledge distillation and advanced learning techniques will be exploited to assist the capability of learning patient representation further.*

## 1. Introduction

In Electronic Health Records (EHRs), information regarding patient status is extensively documented. Therefore, EHR data provides a feasible mechanism to track patient health information and to make better decisions based on data-driven technologies. Unlike data in clinical trials or other biomedical studies, secondary data extracted from EHRs are not designed to answer a specific hypothesis. Instead, their primary goal is to monitor a patient. This results in the issue that EHR data have many challenging characteristics such as *uncurated* (data are not carefully chosen and thoughtfully organized or presented), *poor-quality* (data are rarely subject to data quality audits), *high-dimensional* (thousands of distinct medical events), *sparse* (lots of zero values), *heterogeneous* (drawn from different resources), *temporal* (data are collected over time), *incomplete* (missing values), *large-scale* (a large volume of data), and *multimodal* (multiple data modalities). A wide variety of studies have conducted predictive modeling of EHR data, which is a machine learning task that applies EHR data to construct a statistical model for the purpose of predicting a given clinical outcome of interest [1]. However, the complexity of EHR data discussed as above makes it difficult to directly use EHR raw data in machine learning models to achieve predictive modeling.

A critical element in predictive modeling of EHR data is to effectively convert patient data from the raw EHR format to a machine learning representation—in other words, to transform patient data to meaningful information that can be further understood algorithmically. The effectiveness of predictive models for improving disease diagnosis, phenotyping, and prognosis heavily depends on the quality of this feature representation. In machine learning, the task of representation learning is to learn and extract good feature representations from raw data automatically [2]. Patient representation learning is one particularly promising direction of combining representation learning and large EHR datasets. It refers to learning a mathematical description of patient data to find an appropriate way of transforming raw data into meaningful features. The patient representations built from many EHR data modalities (including clinical narratives, lab tests, treatments, etc.) should be organized in a form that enables machine learning to learn effective prediction models for many tasks.

This work investigates the methods of representation learning and the field of patient representation learning from EHRs through a methodological review of the literature. We collected data on 28 patient representation variables from 49 papers published in a diverse variety of venues, published up to December 2019.

We seek to understand the following research questions:

1. *Resources*: What are the resources available for learning patient representations? How is patient data transformed from raw input to important features?

2. *Methods*: What representation learning methods are being contributed? What kind of models and algorithms are used to develop patient representations?

3. *Applications*: What types of clinical problems and outcomes are addressed?

4. *Potential*: How could these methods potentially contribute to diverse research communities?

## 2. Background

### 2.1. Patient Learning Data Pipeline

First, we briefly summarize the methods and resources that are commonly used in related studies (**Figure 1**). The learning starts with raw patient data from either structured codes such as ICD-9, or unstructured data like clinical notes. After some initial embedding techniques to transform patient data into input features, deep learning models are utilized to create the patient representation. The learning model can be unsupervised, supervised, or self-supervised. The final step is to evaluate the learned representations by applying it to some clinical task(s), such as mortality, readmission, or a specific disease prediction. There are also some ways to visualize patient representations to provide some intuitive understanding or interpretation of the representation. The majority of studies in this field follow this pipeline to learn patient representations from EHR data and further evaluate on downstream tasks.

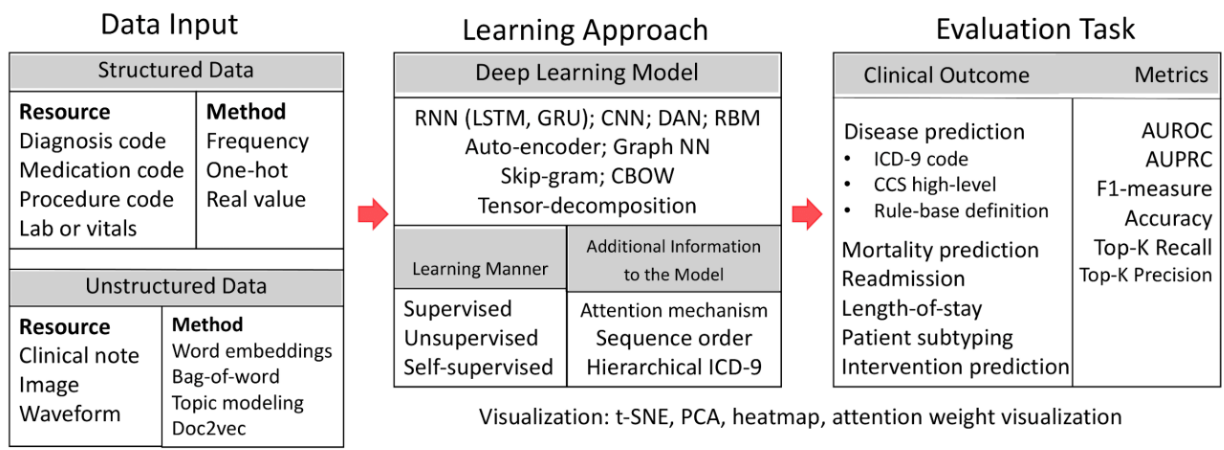

Figure 1. Workflow of patient representation learning

## 2.2. Patient Representation Methods

According to Bengio et al. [2], deep learning models are representation learning methods with multiple layers, each with their own representations. The model starts with raw input, constructs non-linear modules that transform the representation at each layer into high-level, abstract representations. As already discussed, EHR data is high-dimensional and sparse. Thus, deep learning models are particularly well-suited to encode EHR data to learn patient representations. In this section, we will introduce the background of representation learning methods and how they are applied to learn patient representations. We describe five methods of representing a patient: vector-based, temporal matrix-based, graph-based, sequence-based, and tensor-based representations. For more detailed and technical interpretations of deep learning models, we recommend the review by LeCun et al. [3].

### 2.2.1. Vector-based Patient Representation

In vector-based patient representation, every patient is represented by a mathematical vector. Models that attempt to construct vector-based patient representations are as follows:

**Fully connected deep neural network** (fully connected DNN), originally inspired by biological neural networks, contain one or more hidden layers connected in a feed-forward manner [4]. Fully connected DNN are also known as multi-layer perceptrons (MLPs) or dense neural networks. Like the other neural network methods described below, they use backpropagation for supervised learning with non-linear activations. Some early attempts for learning patient representation with neural networks have applied fully connected DNN approaches [5]. However, the majority of recent studies have considered this architecture as a baseline method [6].

**Autoencoders** are an unsupervised deep learning model that learns abstract representations from high-dimensional data, which is also a way of performing dimensionality reduction. An autoencoder learns to predict its input, but must compress the input signal through progressively smaller intermediate layers. The inner layer can be used as a low-dimensional representation. The variants of autoencoders include denoising [7], stacked denoising [8], variational [9], and contractive autoencoder [10]. Given the prior success of autoencoders for representation learning, their application to patient representation learning was first introduced by Miotto et al. [11], where a dense and general patient representation was derived with unsupervised three-layer stacked denoising autoencoders.

**Convolutional Neural Networks** (CNNs) were specifically developed for image processing [12]. Each module of the CNN contains a convolutional layer and a pooling layer. Those modules are stacked to construct a deep CNN. The convolutional layer uses a small neural network that slides across the image, allowing for the capture of local image properties irrespective of their location in the full image. However, CNNs have been applied to more data types than just images, including text [13] and waveforms [14]. More advanced variants of CNN also attempt to incorporate temporal information in addition to the convolutional layer to model longitudinal patient data [15]. This is different from the variants of CNN in image processing fields such as VGGNet [16] and ResNet [17], which add more complex layers in the architecture.

**Word2vec** originated from natural language processing to learn word embeddings from large-scale text resources [18]. There are two algorithms in word2vec: continuous bag of words (CBOW) predicts a target word given the surrounding context, and skip-gram predicts the surrounding context given a target word [19]. Both algorithms have just one hidden layer; thus, word2vec techniques are shallow networks compared with other deep learning methods. The variants of word2vec have been applied to learn representations from clinical codes [20], characterized by different assumptions and aimed to capture relationships between code sequences.

### 2.2.2. Temporal Matrix-based Patient Representation

Temporal matrix-based patient representation constructs a two-dimensional matrix with one dimension related to time and the other dimension related to clinical events from the EHR. Nonnegative matrix factorization (NMF) is an algorithm for decomposing high-dimensional data from a set of nonnegative elements. NMF has been widely used in bioinformatics for clustering sources of variations [21,22]. There were also early attempts of applying NMF or its variants into EHR patient data representation [23–25,15], which are some of the earliest examples of constructing a mathematical representation for each patient. They also showed the challenges of using EHR data and demonstrated the feasibility of encoding the latent factors of temporal patient data by providing a one-to-one identifiable mapping between the patient matrix and the target label.

### 2.2.3. Graph-based Patient Representation

Graph-based patient representation constructs a compact graph for each patient where the nodes in the graph encode clinical events and the edges between the nodes encode relationships among the clinical events. One of the early works developing graph-based EHR representation was proposed by Liu et al. [26], where they designed a novel graph representation algorithm to learn distinct clinical events from EHR data that included temporal relationships among the events. Although their work was not developed based on deep learning, one such emerging application of deep learning into graph representation is the Graph Neural Network.

**Graph Neural Networks** consist of a finite number of nodes and edges to connect data. Nodes contain information about entities, and edges contain relations between entities. Prominent methods of GNN include Directed Acyclic Graph [27], Graph Convolutional Network [28], Graph Attention Network [29], and node2vec embeddings [30]. GNNs attempt to learn graphical structures of EHR data that can infer the missing information through other representation mechanisms, therefore resulting in a more explainable representation. GNNs are particularly useful to introduce domain knowledge into the architecture. For instance, a few studies have employed hierarchical ICD-9 knowledge graphs as the graph model to enhance the interpretability and performance of the proposed method [27,31]. Other knowledge-based resources include adverse drug-drug interactions [32], and comorbidities [33].

### 2.2.4. Sequence-based Patient Representation

Sequence-based patient representation produces a timestamped event sequential features for each patient.

**Recurrent Neural Networks** (RNNs) were developed for processing sequential inputs such as language [34,35]. RNNs deal with a sequence of inputs one item at a time and transfer the hidden state of each input unit to the next input unit, so the current state implicitly contains information about the entire sequence history. The common variants of RNN include Gated Recurrent Unit (GRU) and Long Short-Term Memory (LSTM), both designed to minimize the vanishing gradient problem. GRU adds a gating mechanism into the RNN [36]. LSTM is particularly useful for dealing with long-term dependencies [37]. Thanks to the advances in RNN architecture, many studies have applied RNNs to develop patient representations where they have focused on representing patients using combinations or sequences of clinical codes. One such RNN-inspired approach is Doctor AI [38], where distributed vector representations of clinical codes are derived to represent patient trajectories.

Although RNNs enable to sequential modeling, especially some variants of RNNs like LSTMs are effective in dealing with long dependencies, limitations of RNNs still persist. One notable limitation is that RNNs cannot be trained in parallel (which will increase training time). Also, RNNs only process information from one direction, and even bi-directional RNNs which encode from two directions are a simple concatenation of two directions. Therefore, the **Transformer** architecture equipped with self-attention and positional embeddings was introduced to achieve true bi-directional representation [39].

Recently, with the wide adoption of Transformer-based language models in NLP (e.g., GPT [40], BERT [41]), Transformers continue to be applied for patient representation learning and for clinical sequence modeling [42–45]. Similar to the use of RNNs to enable sequence modeling of EHR data, Transformers also encode each clinical event at every time stamp as a unit and an entire patient trajectory as a whole sequence. Unlike a RNN using recurrence to predict the next unit, the Transformer architecture considers the sequence of units as a whole and employs self-attention to learn the essential information from the entire sequence.

### 2.2.5. Tensor-based Patient Representation

Tensor-based patient representation is a method that represents each patient with a three-dimensional (or more) tensor consisting of events such as diagnoses and treatments.

**Tensor Decomposition** is the high-order extension of matrix decomposition that seeks to decompose high-dimensional tensors into products of low-dimensional factors [46]. The CANDECOMP/PARAFAC alternating Poisson regression (CP-APR) algorithm is one of the representative methods that are well-studied for tensor decomposition, trying to express a tensor as the sum of a finite number of rank-one tensors [47]. Compared with traditional dimensionality reduction methods, tensor decomposition is unique in having the capability of managing multi-aspect features in multiple dimensions and is versatile enough to incorporate domain knowledge into the operation [48]. These advantages have made tensor decomposition a promising modeling approach to learn abstract patient representations from EHR data and provide good interpretability and scalability [49]. Each tensor is constructed of patient-level data, with three different factors including a patient factor, and two other factors related to clinical events (i.e., diagnosis with medication, or diagnosis with procedure). Each tensor constitutes a phenotype with a weighted sum of rank-one tensor from the outer product of three factor vectors. A tensor-based patient representation allows for the capture of complex interactions and relationships between clinical events (especially phenotypes, comorbidities, and medications) that are not evident in flattened EHR data [50].

### 3. Materials and Methods

Papers that are eligible for inclusion in this review are characterized by (a) focus on patient representation learning, (b) use of patient data from longitudinal EHRs, and (c) utilization of deep learning or neural network models. Notably, we consider studies using deep learning as one criterion because the unique characteristics of EHR data (as discussed in Introduction) make deep learning a feasible data-driven solution to generate robust representation simultaneously from diverse resources compared with other techniques. It is likely that, at least in the next few years, most advances in patient representation will utilize deep learning approaches.

These criteria were applied to generate keyword searching queries for literature database search. Five databases consisting of PubMed, EMBASE, Scopus, the Association for Computing Machinery (ACM) Digital Library, and the Institute of Electrical and Electronics Engineers (IEEE) Xplore Digital Library were queried on December 31, 2019. The search consists of the following combination of keywords:

> *("deep learning" OR "neural network") AND patient AND (vector OR representation OR embedding OR vec) AND ("electronic health records" OR ehr OR "electronic medical records" OR emr).*

As for publication characteristics, we limit our search to English-language research articles with the publication time range from 2000 to 2019 and only focus on peer-reviewed journals and conference proceedings (i.e., posters and preprints are not included). For study characteristics, only study designs related to developing a deep learning- or neural network-based patient representation from electronic health records are considered. The study outcome should also apply the learned representation to downstream evaluations of clinical predictions. In terms of exclusion criteria, studies such as review papers and proceeding summaries are excluded. Studies that use patient data other than EHRs are also excluded.

Duplicates were removed and additional records were added using a snowballing strategy. The additional records were subsequently traced from references of the included papers and personal readings. Next, the articles were uploaded to Rayyan [51] and screened for title and abstract. Two authors (YS, JD) screened the articles under a blind format. At the full-text screening step performed on Zotero [52], each study was read in full by one author (YS) to validate the final relevancy.

As a result, 363 papers were queried at the beginning, including 349 articles retrieved from the five databases, and 14 additional records identified through snowballing. At deduplication, 54 duplicate records were removed and 309 records remained. At the step of title and abstract screening, 199 records were excluded because the topic is not relevant (n=114), they are traditional predictive modeling of the single task (28), they are not peer-reviewed or original research papers (26 studies are not research, 5 preprints), no deep learning is used (12), no evaluation of the patient representations (7), and there is no EHR data (7).

After these exclusions, 110 papers went through to the next screening step (full text screening) to determine the relevancy of each study. At this step, 61 papers were excluded due to no patient representation being developed (n=37), no deep learning method being used (8), no EHR data being used (7), no evaluation (6), and only containing abstracts (3). Finally, this left 49 articles to be included in this work. A detailed literature selection procedure (PRISMA diagram) [53] is shown in Figure 2. Based on the variables extracted from 49 publications, some interesting findings are worthy of being highlighted and discussed in the following sections.

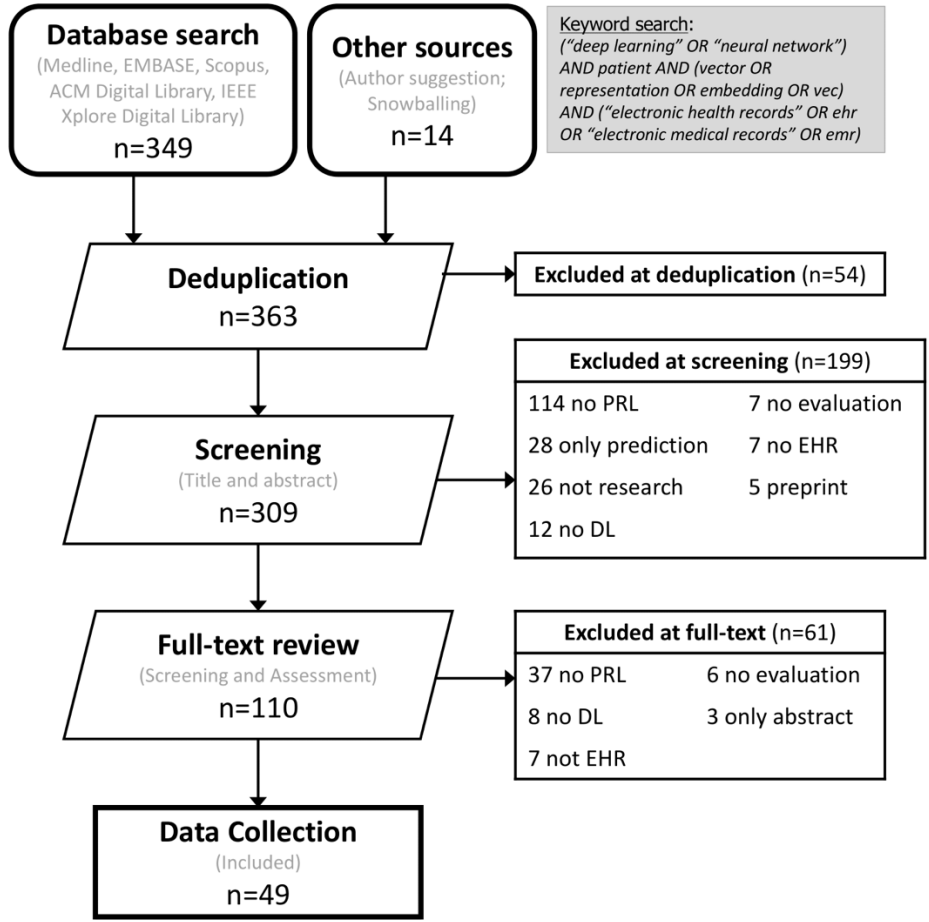

Figure 2. PRISMA flowchart for including articles in this review

(PRL: Patient Representation Learning. DL: Deep Learning. EHR: Electronic Health Records.)

We performed data extraction from the final included papers considering the following variables:

- Publication date: the date on which this paper was published by the conference or journal.

- Publication venue: the conference or journal where this paper was published.

- Representation learning approach: this contains two items including the proposed deep learning model(s) and additional attributes to enhance the model.

- Patient data type: what modality of EHR data was used? For instance, structured code from diagnoses/procedures/medications tables, numerical values from lab measurements, or unstructured data from clinical notes/images/waveforms.

- EHR resources: where does the patient EHR data came from? For example, publicly available datasets or a private local hospital data warehouse? What is the size/duration?

- Preprocessing methods: what is the format of the input data that was fed into the model architecture? In other words, how the EHR raw data was preprocessed into numerical data?

- Clinical outcome tasks: main clinical tasks for evaluating learned patient representation.

- Evaluation metrics: quantitative measure of the performance of prediction/clustering/regression on clinical outcome tasks.

- Interpretability: the degree to which a human can understand the model's result or learned patient representations.

- Objective functions: the loss function when optimizing the models.

- Computational resources: for instance, deep learning frameworks and computational platforms (GPU).

- Code reproducibility: reproducibility of the results/instructions on open-source platforms.

## 4. Results

### 4.1. Publication Characteristics

First, we look at the publication numbers over time in **Figure 3**. The volume of studies is currently growing rapidly each year (almost 3 times in 2017 over 2016). Although there is a slight decrease from 2018 to 2019, we think this is because we exclude preprints at the first step of screening. Apart from the publication date, we also outline the major research communities that have contributed to the topic. **Table 1** lists the top conferences or journals that have published this topic in three main communities. The conferences or journals include, but are not limited to, AMIA Symposium, AMIA Joint Summit, Conference on Information and Knowledge Management (CIKM), Journal of the American Medical Informatics Association (JAMIA), Journal of Biomedical Informatics (JBI), BMC Medicine, Scientific Data, Scientific Reports, NPJ Digital Medicine, Neural Information Processing Systems (NeurIPS), Knowledge Discovery and Data Mining (KDD), Machine Learning for Health Care (NeurIPS_ML4H), and Association for the Advancement of Artificial Intelligence (AAAI).

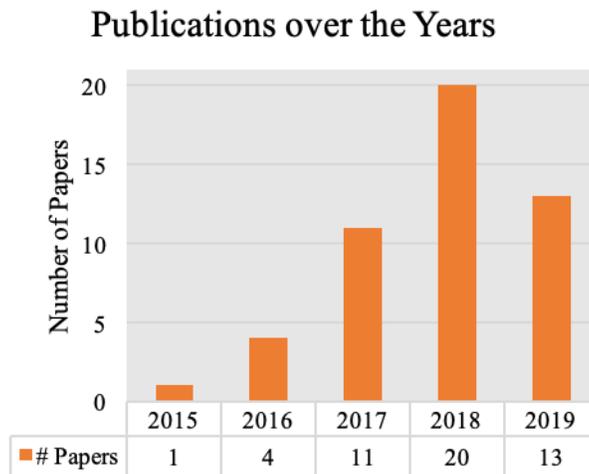

Figure 3. Publication counts over time.

Table 1. Selection of conferences/journals publishing the included studies.

| Community | Conference/ Journal | Name (# Papers) |
|---|---|---|
| Informatics | Conference | AMIA Summit (1),  AMIA Symposium (1), others (4) |
| | Journal | JBI (3), JAMIA (1) |
| Medicine | Journal | NPJ Digital Medicine (1), Scientific Reports (2), PLoS ONE (1) |
| Computer Science | Conference | KDD (8),  AAAI (4),  CIKM (3),  NeurIPS_ML4H (4),  IEEE-related (7), others (7) |
| | Journal | JMLR (2) |

## 4.2. Study Characteristics

Pertinent study characteristics of the 49 reviewed publications are reported in the following sub-sections, and also summarized in **Supplementary Table 1**. These study characteristics enable qualifying the scientific validity and applicability of each study, which reveals the inadequacies and gaps of current research fields, and would be potentially useful to inform the need for new study designs for future research in patient representation learning.

### 4.2.1. Learning Scenarios

We find that, among the 49 included studies, the majority (n=30) applied supervised approaches to learning the representation. Here one would train and optimize the representation with some target or objective, usually focusing on phenotyping or other clinical outcomes. Unsupervised learning was applied in 11 studies. Unlike supervised learning, unsupervised learning does not require labels. This would include the use of autoencoders, which function by reconstructing the data from its raw input. In all forms of unsupervised learning, patient representations are learned separately from the prediction task. Since no labels are required, unsupervised learning is more likely to benefit from large EHR datasets.

Self-supervised methods were used by 8 studies. Unlike supervised learning that relies on labels for a target task, or unsupervised learning that has no label-driven objectives, self-supervised learning frames a learning scenario as predicting a subset of information using the remaining data. Essentially this enables the use of models that are commonly used for supervised tasks (including many deep learning models) in a setting that does not have the labels of interest for the target task. With self-supervised learning, the patient representation is obtained similar to word embeddings in NLP studies, where a target word is predicted with its contextual information. Since patient representations are themselves largely intended as intermediate steps to a final prediction task (usually done via supervised learning), learning the representations in a self-supervised manner provides many of the benefits of unsupervised learning (no manual curation of labels) while still allowing for the use of powerful deep learning methods (e.g., transformers) that can otherwise be considered supervised. The self-supervised tasks included in this review are mainly used for clinical code representation learning, where clinical codes are ordered sequentially, and the codes in between are masked and predicted using the context of clinical codes nearby, similar to the algorithms of skip-gram and CBOW used to learn word embeddings [19].

### 4.2.2. Patient Data Types

In terms of patient data types in EHRs, among the 49 studies, 37 papers used structured codes (i.e., diagnosis codes, procedure codes, medication codes) to build the representation, while 6 papers used unstructured notes. The remaining 6 papers used both structured codes and unstructured notes jointly. Though combining heterogeneous resources is promising, we noticed these studies were not really combining the raw data fully from the two modalities into the same model. Instead, they only used a subset of data from one modality [6,11,54–57]. One such method is to simplify clinical notes by using topic models to represent unstructured resources [6,11,55].

### 4.2.3. Preprocessing Methods

Although not many studies emphasized the preprocessing steps used to transfer raw input data into the model, we assume this is because deep learning does not necessarily need heavy manual work for preprocessing. However, we still extract the preprocessing approach and workflow of each study because we consider this an essential step towards representing a patient. Among 37 studies learning from structured codes, the majority (n=27) generate one-hot vectors for the input data, and the frequency of a given code is used in 5 studies [11,48,58–60]. For inputs from multiple resources, input vectors are concatenated before feeding them into the architecture [14]. Normalized formats of patient data are also important to learn an effective representation. For instance, z-score is applied to convert data from continuous variables to discrete one-hot vectors [55,57,61]. For both structured codes and unstructured notes, standardized vocabularies including FHIR [6] and UMLS CUIs [62,63] are introduced to transfer raw tokens into unique phrases to reduce the sparsity of the input data.

### 4.2.4. Learning Models and Architectures

A primary interest of this study is the types of deep learning models that the proposed methods used as the foundation to learn patient representations. We plot the number of papers that have applied each deep learning model as the foundation architecture, and also differentiate studies with different data input types, shown in **Figure 4**. In total, there are 13 studies (structured: 8; unstructured: 1; combined: 4) that have applied an LSTM model to develop patient representations, which is the most common architecture. CNNs are the next with 11 studies (structured: 7; unstructured: 2; combined: 2). GRU is also common with 11 studies (all structured EHR data). This result is reasonable since both CNN and RNN architectures are utilized to learn local or sequential information from large-scale data and to find the nuance of variance within the data.

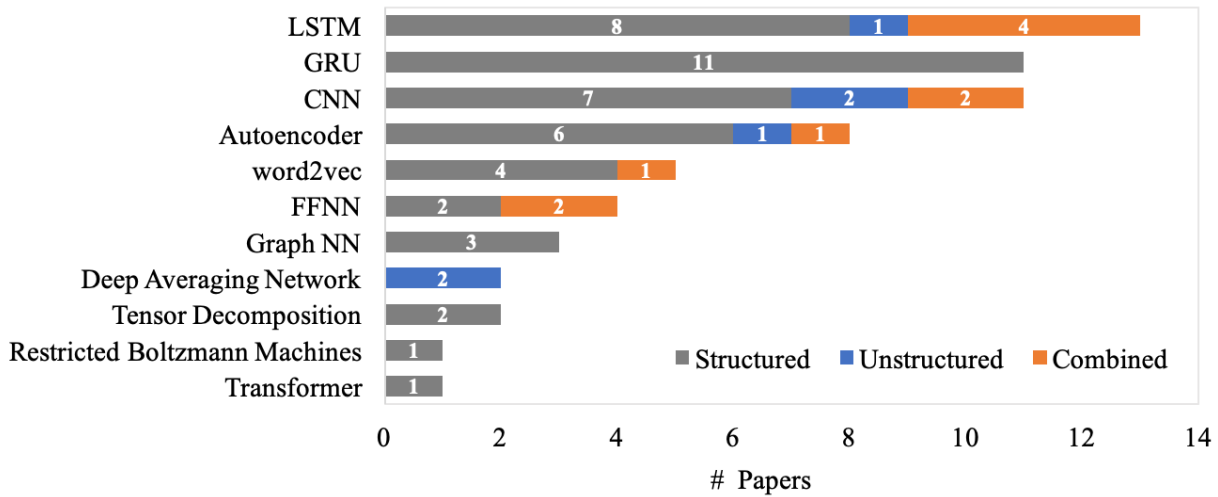

Figure 4. Deep learning architectures used in the included studies

Table 2. Correspondence between patient data types and deep learning models

| | Patient Input Data Type | | |
|---|---|---|---|
| **DL Models** | **Structured** | **Unstructured** | **Combined** |
| LSTM | [14,58,61,64 – 68] | [69] | [6,54,55,57] |
| CNN | [14,33,66,68,70–72] | [63,73] | [54,55] |
| GRU | [31,38,72,74–81] | | |
| Autoencoder | [61,67,74,75,82,83] | [84] | [11] |
| word2vec | [59,85–87] | | [56] |
| FNN | [88,89] | | [6,54] |
| Tensor Decomposition | [48,58] | | |
| Deep Averaging Network | | [62,90] | |
| Graph NN | [27,31 – 33,91] | | |
| Restricted Boltzmann Machines | [60] | | |
| Transformer | [43] | | |

Although the methods of learning patient representations are all built upon deep learning models, many unique architectures were introduced, and additional learning modules have been applied to capture the characteristics of EHR data. To improve performance and provide a degree of interpretability to the model, an attention mechanism is adopted in 13 studies. A hierarchical knowledge graph has been incorporated into learning representations in 6 studies [27,31–33,81,91].

Sequential order between clinical events has been considered in the architecture in 27 studies. The majority of them have applied RNN (i.e., LSTM, GRU) to model sequential order with time-series patient data from diagnoses or measurements. Specifically, RNN adopts hidden states to carry information from the previous step to the next, which attempts to transmit sequential order between clinical events. Clinical events at each timestamp (i.e., diagnoses at each visit, or lab measures at every hour) are fed into every unit of the RNN, and the hidden state at this time point is updated with the hidden states from the previous timestamp and additional input from the current timestamp. Along these lines, sequential information from discrete event time-series can be transmitted to the final prediction target. The remaining 22 studies have not considered sequential information, based on varied ways of preprocessing patient data. For instance, deep learning models that fail to process sequential orders such as CNNs or autoencoders are applied in the architecture [73, 98], the patient EHR data are only derived from clinical notes [62, 90], and the patient EHR data are represented with one-hot encoding or frequency counts [11].

An interesting observation when speculating about what the "ideal" architecture for learning patient representations would be is that deep learning models are not mutually exclusive from each other. The best aspects of each model can be combined into a joint deep learning architecture. For instance, RNNs can assist with autoencoders to learn unsupervised representations while improving their ability to capture sequential clinical events [61,67,74,75]. CNNs can also be combined with RNNs to learn multifaceted patient representations [14,55,72].

### 4.2.5. Patient Data Resource

We have observed the recent success in the image processing field built upon the curation of a very large image dataset, ImageNet [92]. In this review, we investigated the resources of available EHR data that are large-scale and publicly available to advance research. Although it may be hard to access EHR data due to privacy concerns, a large publicly available clinical dataset can push representation performance to a scale similar to that of image processing by giving researchers a common benchmark to build on prior work. We observed that the prominent dataset in 49 included studies is the Medical Information Mart for Intensive Care (MIMIC-III) [93], where 22 out of 49 studies have used this dataset. Because MIMIC-III has both structured and unstructured data, clinical outcomes studied using MIMIC-III are widely varied, including mortality prediction, disease diagnoses, phenotype predictions, readmission, and length-of-stay forecasting. Other publicly available datasets that have been used to develop patient representations are: the Parkinson Progression Marker Initiative (PPMI) [67,74,94], the Alzheimer's disease neuroimaging initiative (ADNI) [74,95], the i2b2 obesity challenge [62,63,90,96], and eICU Collaborative Database [42,97]. We summarize the public datasets and their corresponding clinical applications in **Table 3**. Considering the public datasets for each clinical task, the most popular is MIMIC-III for disease predictions (n=10). Private datasets from local clinical data warehouses were used in 27 of the 49 studies.

The detailed description of local hospital datasets can be found in **Supplemental Table 2**. A total of 9 studies have applied their methods on two or more datasets (i.e., private or public).

Table 3. Benchmark dataset and clinical outcome applications

| Public Resource | Description | Patient Data Type | Patient Count | Application Tasks | Paper |
|---|---|---|---|---|---|
| MIMIC-III | A critical care database, including de-identified patient data with ~60,000 Intensive Care Unit (ICU) admissions [93] | Structured; Unstructured | ~46,000 | Mortality Prediction | [14,43,57,58,61, 65,73,77] |
| | | | | Disease Prediction | [27,31,43,54,56, 57,59,91,98,99] |
| | | | | Admission Prediction | [77,88] |
| | | | | Length-of-stay Prediction | [14,43,73] |
| | | | | Patient Similarity | [58] |
| | | | | 24-hour decompensation | [14,43,61] |
| | | | | Intervention Prediction (ventilation, prescription, lab test order, etc.) | [32,54,55,65,98] |
| PPMI | A longitudinal patient dataset comprising clinical and behavioral variables, imaging, and specimen data of Parkinson's disease patients. [94] | Structured | ~1,000 | Patient subtyping | [67,74] |
| ADNI | A longitudinal patient dataset consisting of assessments collected from selected patients in varied stages of Alzheimer's Disease. [95] | Structured | ~600 | Patient subtyping | [74] |
| i2b2 obesity challenge | 1237 discharge summaries from the Partners HealthCare Research Patient Data Repository. Each discharge summary was annotated with patient disease status corresponding to obesity and fifteen comorbidities of obesity. [96] | Unstructured | ~1,000 | Phenotype Prediction | [62,63,90] |
| eICU Collaborative Database | A combination of multiple critical care units across the United States. The data covers patients who were admitted to critical care units in 2014 and 2015. [97] | Structured | ~139,000 | Mortality medication, and diagnosis predictions | [42] |

MIMIC-III: https://mimic.physionet.org/
PPMI: http://www.ppmi-info.org/
ADNI:  http://ida.loni.usc.edu/
eICU: https://eicu-crd.mit.edu/about/eicu/

**4.2.6. Clinical Outcomes and Applications**

Studies involving patient representation learning evaluate their representations based on risk predictions. The general assumption is that a more advanced and robust patient representation would improve the performance of predictive models. **Table 4** illustrates the clinical outcomes addressed among the 49 papers. We categorize main clinical tasks for each paper into seven types: disease prediction (n=31), mortality prediction (12), length-of-stay forecasting (8), admission prediction (7), patient subtyping (4), intervention prediction (5), and medical cost forecasting (3). To show the generalizability of the study, the majority of works have applied two or more clinical outcomes to evaluate the representations, resulting in a number of studies that have sizable overlap in terms of outcomes studied. The most common clinical tasks (31 studies) are disease diagnosis predictions, which is to predict whether the patient will develop a given condition. Mortality prediction is the second most common task with 12 studies.

Rather than focusing on predictive tasks, patient representations can be potentially useful for clustering patients based on similarity, which can be used to identify patient subtypes, among other things. Patient subtyping is helpful for studying certain types of diseases, including Parkinson's Disease. However, very few studies have focused on patient subtyping (n=4). We assume this is because patient subtyping is not like other prediction tasks such as mortality and disease predictions that have straightforward label information. Typically, a data-driven technique is required to differentiate between different subtypes. For instance, the technique of measuring the distance of two patient representation vectors is to calculate the similarity; thus, to decide whether the two patients belong to the same subtype. Zhang et al. [100] proposed a data-driven subtyping method with LSTM and Dynamic Time Warping to first transform patients into temporal representations and then calculate the similarity of every patient pair.

Evaluation metrics are mainly dependent on the prediction task. We summarize the metrics for the above clinical outcomes in **Table 4**. Typically, for prediction tasks, AUROC, AUPRC, Precision, Recall, F1-measure, and Accuracy are the most used. Metrics including Top_K_Recall or Top_K_Precision (ranking metrics from information retrieval), are adopted particularly for predicting full sets of diagnoses where K is the number of diseases. In addition, for regression tasks that involve predicting a real-valued quantity (i.e., medical cost, readmission forecasting), Root Mean Squared Error (RMSE), Mean Absolute Percentage Error (MAPE) and Goodness-of-Fit R-squared are utilized.

Another evident observation stands out when we look into the time periods for the evaluation, that is how to select appropriate observation and prediction periods. Normally, clinical events in the observation period are accumulated to construct the samples for training, while clinical events in the prediction period are considered as the gold standard for evaluation. We discover an interesting connection that the design of observation and prediction periods are highly correlated with the clinical outcome tasks that are evaluated. **Table 4** shows the time period for prediction with corresponding clinical tasks. Prediction tasks, such as disease, readmission, patient subtyping, and medical cost forecasting, generally extend observation windows for long time periods. Month-level [69, 88], year-level [11, 48, 66, 78, 80], and visit-level predictions [31, 38, 56, 59, 64, 70, 71, 76] are prevalent for those tasks. Take disease prediction as an example, a number of studies extracted diagnoses from the first visit and then predicted diagnoses in the next visit (i.e., visit-level observation) [31, 38, 56, 59, 64, 70, 71, 76]. On the contrary, for those outcomes that happen in hospital, such as inpatient mortality [14, 57, 61, 65], length-of-stay [14, 83], and intervention prediction [32, 55, 65], observation windows at hour level are developed. Specifically, 24 hours are common as these tasks are immediate and physicians make decisions instantly. Rajkomar et al., [6] assessed two periods for mortality prediction (i.e.,12 hour and 24 hour) and showed that, if the deep learning model and other parameters remained the same, the prediction made 24 hours after admission attained a slightly better performance than that of 12 hours. This outcome is reasonable because more training features were accumulated with a longer time of observation. However, we assume this is a trade-off between time and accuracy, because the decision for this type of demanding task should be up-to-date as well.

Table 4. Clinical outcome of interests with evaluation metric(s) and prediction period level

| Evaluation Tasks (# paper) | Evaluation Metrics | Time periods | Papers |
|---|---|---|---|
| Disease Prediction (31) | AUROC, AUPRC, Accuracy, Top K Recall/Precision | Year-level, visit-level | [6,11,27,31,33,38,43,54,56,57,59, 60,62–64,66,67,69–72, 75, 76, 80, 81, 83, 86, 89–91, 98] |
| Mortality Prediction (12) | AUROC, AUPRC, F1, Accuracy | 24-hour, month-level | [6,14,43,57,58,61,65,73,75,77,85,98] |
| Length-of-stay Prediction | AUROC, AUPRC, F1, Accuracy, MSE, MAPE | 12-hour, 24-hour | [6,14,43,73,82,83,85,87] |
| Admission Prediction (7) | AUROC, AUPRC, Accuracy, Top K Recall/Precision | Month-level, year-level | [6,38,48,77–79,88] |
| Patient Subtyping (4) | Significant difference among different groups; NMI, MSE, Sparsity and Similarity | Visit-level | [67,70,74,79] |
| Intervention Prediction (5) | Jaccard, Recall, Precision, F1 | 24-hour | [32,54,55,65,98] |
| Medical Cost (3) | R-squared, RMSE | Year-level | [48,85,87] |

### 4.2.7. Objective Function

Only a few papers have reported the loss function applied in the study. We find that cross-entropy loss (or logarithmic loss) was the most common objective function for optimization. Aside from traditional cross-entropy loss, a weighted loss function was applied to account for an imbalanced-class dataset [55]. For regression tasks, mean squared error (MSE) loss and mean absolute error (MAE) loss were mainly performed [43]. What's more, studies developing patient representations with several learning objectives conduct a unified loss function with different targets being optimized jointly. For instance, a masked binary cross-entropy loss was employed for a multi-task architecture [43,69]. Doctor AI [38] has a joint loss function containing the cross-entropy loss from diagnosis predictions and the squared loss for the prediction of time duration for a single patient.

### 4.2.8. Interpretability

While patient representations may be relatively low-dimensional (often less than 500 dimensions), these collections of continuous numbers can be particularly abstruse and difficult for humans to interpret [101]. Visualization techniques are often performed to construct qualitative clusters of patient cohorts to provide an intuitive interpretation of the learned representations [102]. Such techniques for patient representations include t-Distributed Stochastic Neighbor Embedding (t-SNE) [103], Multidimensional Scaling (MDS) [104], Principal Component Analysis (PCA) [105], and Uniform Manifold Approximation and Projection (UMAP) [106]. Among these four methods, t-SNE has developed into the standard tool in this field due to its capability of revealing clusters in data. Ideally, patients typically from test sets associated with different labels, are projected and clustered into distinct groups. We observed 8 papers visualized with t-SNE to interpret the representations [14,27,31,60,66,73,75,79], and one study visualized with MDS [70]. Apart from projecting representations, interpretability can be provided by visualization methods including a heatmap of features [33,67,78], visualizing attention weights [57,72,76,80,88], and case study (explanation by examples) [32].

### 4.2.9. Computational Resources and Reproducibility

With the rapid improvements in computational resources, GPU-accelerated techniques have been exploited to train deep learning models faster and more efficiently. Because the literature in this study is from recent years (2015-2019), we observe the majority (31 out of 49) have reported the computational resources along with the deep learning frameworks. The reported deep learning frameworks include Theano (n=12), TensorFlow (9), PyTorch (6), and Keras (4).

The need for reproducibility of scientific results has been growing in awareness in biomedicine in recent years. This is also true in this field of research because we want to ensure that the patient representations are generalizable and reproducible across multiple institutions. Unfortunately, reproducibility in patient representation learning is difficult due to the variety of data resources, learning methods and architectures, and data preprocessing steps. Another major difficulty in data sharing is related to privacy concerns as inappropriate access to sensitive patient data might lead to patients' identity disclosure. While there are machine learning techniques that preserve data privacy [107–111], it is often impossible to share patient data due to privacy issues. Thus, providing workflow and computer code associated with publications is becoming increasingly common to enable the reproducibility of the method, if not the full work. One such open-source platform, GitHub, manages a robust infrastructure that is appropriate for sharing the full suite of intermediate experiments such as programming code, statistical results, etc. Therefore, we consider papers that provide code as one measurement of reproducibility. Among the 49 included papers, 20 of them have provided code links. Notably, the source code of RETAIN [80], based on Theano, has been starred on Github more than 100 times and forked over 50 times, and it has been redeveloped in other deep learning frameworks (TensorFlow, Keras). RETAIN was also commonly used as baselines in the subsequent series of advanced models (i.e., GRAM [27], Dipole [76], Health-ATM [72]). Thus, we anticipate that sharing open-source resources and code would motivate researchers to further explore innovations in a more scientific manner.

## 5. Discussion

### 5.A. Growing Importance

In this review, we provide an overview of the current research into EHR patient representation learning. A total of 363 articles were assessed, with 49 studies from 2015-2019 meeting the full criteria for review. We observe a growing trend of developing deep learning-based patient representations from EHRs. The earliest attempt of learning patient representations with deep learning used Restricted Boltzmann Machines, proposed in 2015 [60]. An increasing number of works continued to contribute to improving the representation power of deep learning-derived representations.

Notable observations are identified throughout the review and can be addressed to the four questions we proposed at the end of Introduction.

1. Resources: We investigated that the resources of available EHR data are quite limited due to privacy concerns. The most prominent dataset is the Medical Information Mart for Intensive Care (MIMIC-III), where 22 out of 49 studies used this dataset.

2. Methods: Learning was mainly performed in a supervised manner optimized with cross-entropy loss. Recurrent Neural Networks were the most common deep learning model.

3. Applications: Disease prediction was the most common application and evaluation. We also identified a wide variety of clinical outcomes being addressed, including mortality prediction, admission prediction, length-of-stay forecasting, patient subtyping, intervention prediction, and medical cost forecasting.

4. Potential: Enhancements in patient representation learning techniques will be continuously growing for powering patient-level EHR analyses. We anticipate that researchers from diverse communities will leverage the richness and potential of EHR data, and will assist the capability of learning patient representation further.

Furthermore, analyses and evidence from data extractions of the 49 included papers suggest the importance and feasibility of learning comprehensive representations of patient EHR data. Studies that forego any attempt to utilize patient representations as an intermediate structure lose out on several advantages. Most notably, many representation learning models utilize—through unsupervised or self-supervised learning—very large amounts of EHR data. In contrast, traditional predictive models focus on just the labeled information needed for the single task they are trained on. This is similar in essence to the difference between discriminative and generative machine learning models, where the latter attempt to model the full distribution of the data. Since learned representations attempt to gain a cohesive picture of a patient's data, these models may very well be more robust to small changes in the data distribution (e.g., when porting a model to a new institution without re-training). The extent to which this may be true requires further investigation, but it is worth noting that the field of natural language processing has become dominated recently by transformer-based models [41] that essentially follow this approach. Contextual word representations are learned on large corpora in a self-supervised manner ("pre-training") and then those representations form the basis on which prediction models are built ("fine-tuning").

The capability to develop effective patient representations from EHRs derives primarily from how to model the pertinent characteristics of EHR data. For instance, recent clinical events are more likely to contribute to the final label compared to earlier events [80]. Another challenge is that clinical events may be correlated with each other (i.e., diagnoses lead to treatments, disease co-occurrence) [32,81]. Leveraging this inherent structure is a key factor to improving the representation. This has resulted in a wide range of methods that can complement deep learning models to enhance representation power. These methods include developing additional components to models for capturing a hierarchical representation from event-level, to visit-level, to patient-level [81,112]; for encoding the longitudinal factors between clinical events [65,77]; for incorporating domain knowledge into the representation [27,31–33,91]; and for adopting multiple modalities of EHR data [11,14,57].

## 5.B.  Future Directions

Building upon current promising trends in learning meaningful patient representations, we believe the growth will continue, and the challenges of EHR data can be addressed in the following directions. We highlight a limited number of works published after the end of the review period (i.e., since January 2020) to exemplify these promising directions.

### 5.B.1.  Methodology

With the pace of deep learning models diffusing into patient representation learning, algorithm and model development continues to enhance in order to adequately consider and model EHR data [113]. Some pilot works have applied advanced methods to model contextualized information into the learning architecture, such as the transformer model [39]. For instance, Choi et al. [42] proposed the Graph Convolutional Transformer (GCT) to learn hidden patterns of EHR data. This work is an advancement over works like MiME [81] in that GCT can capture hidden logical relationships between structured EHR data (e.g., which diagnoses lead to which treatment) that are completely missing in the raw data. There is also a recent NLP study attempting to show the expressive power of transformers in that they can be universal approximators of sequence-to-sequence aggregators with compact support [114], though this study was not in the clinical domain. Another groundbreaking technique fueled by language models highlights the potential to learn generic representations from raw data. Ethan et al. [115] proposed language model-based representations to learn structured data from EHRs. Si et al. [116] applied pre-training and fine-tuning techniques to learn general and transferable clinical language representations. Li et al. [44], inspired from the architecture of BERT, proposed BHERT, a transformer model for structured data in electronic health records. BHERT is scalable to perform well across a wide range of downstream predictions.

Advanced learning techniques including transfer learning, multi-task learning, and meta-learning can combat data scarcity and label insufficiency to some extent. One of the earliest attempts to learn patient representations, Doctor AI [38], also demonstrated the impact of transfer learning from large-scale clinical data (i.e., 263,706 patients in Sutter Health) to tasks with relatively small amounts of data (i.e., 2,695 patients in MIMIC-III). Furthermore, with the success of multi-task learning in the open domain, such as in image processing and NLP, a growing number of studies also highlighted the potential and feasibility of properly integrating multiple related tasks to learn meaningful patient representations [73,89]. The concept of meta-learning is similar to transfer learning in that it takes advantage of information from other related high-resource domains. We have included a recent novel study in this review work, known as MetaPred [66]. MetaPred applied an optimization-based meta learning built on Model Agnostic Meta Learning (MAML) [117] in clinical scenarios. Because insufficient data is an obstacle for machine learning to solve many clinical problems, such as diagnosing rare conditions, we hope to leverage as many related datasets and knowledge resources as possible. Progress of advanced techniques for learning patient representations are responsible to take knowledge base into account to ensure the clinical relevance of problems and solutions [118].

### 5.B.2. Data Sources

Despite the richness and potential of EHR data, few studies take full advantage of the heterogeneity of EHR data. Instead, many only use a subset of the data. For instance, many studies use diagnosis codes to represent patients, and ignore the real-valued measurements associated with the diagnoses, such as lab test values. Unstructured data, additionally, are often overlooked when developing patient representations (only 12 out of 49 studies used unstructured data). Future work will still be devoted to leveraging rich, yet varied, information in EHRs. One technique known as deep learning-based multimodal representation learning can be exploited to narrow the heterogeneity of EHR data among different modalities. Recently, a joint representation learning strategy named HORDE was proposed by Lee et al. [119], which applies several graphical modules to embed different types of data sources dynamically from EHR data.

The cause of many diseases is very complicated. Many factors, including inherited genetics, living environments, daily diets, and habits, have an impact on disease development. Just relying on the phenotype and medical history information in the EHR would generally fail to effectively represent the patient's full health status. Therefore, integrating multi-source information, e.g. genomics and clinical imaging, would be critical for accurate patient representation learning as well as the associated downstream clinical tasks [120]. Although most current work focuses on a single type of data, there are also some pilot works that are trying to harmonize multiple types of data. Cheerla et al. [121] developed a multimodal deep learning method to predict the survival of pan-cancer prognosis using clinical, multi-omics, and histopathology data together. Some researchers are also investigating the linkage of imaging to genetic data [122] and EHR to genetic data [123] for clinical outcome prediction and disease heritability estimation. These works show the enormous potential to integrate multi-source information to effectively model patient health status [22]. All these data sources can expand the set of features that potentially contribute to fine-grained patient representations.

### 5.B.3. Multi-Institutional Data

Ethical and legal issues with regards to the use of EHR patient data for research, including AI, are well known. The problems lie in the trade-off between the privacy of patient data and data access for research. Yet learning representations from multiple institutions is critical to achieving both scalability and generalizability. Privacy-preserving algorithms and constrained domain adaptation settings are developed to compensate for this tradeoff to some degree [124]. Distributed learning and federated learning are such paradigms that enable the safe use of data from multiple institutions by only learning mathematical parameters [125] and avoiding any instance that might trace back to a specific patient. Distributed learning is an algorithm that, in a parallelized fashion, iteratively learns a single model on separate datasets and obtains the shared model as if data were centralized [126]. Federated learning is a machine learning technique to train different models on multiple decentralized datasets, in contrast to traditional machine learning in which all samples are combined [127,128]. They both attempt to collaboratively train while keeping all the training data at their original institutions. The difference between the two learning paradigms is that distributed learning trains a separate model for each split of the data, and each separate model is transferred to a centralized model, while federated learning trains individual models on heterogeneous datasets. Many existing studies have exploited distributed learning and federated learning across multiple institutions in fields such as medical image processing [129,130], clinical decision support [131], and clinical oncology [132], to facilitate privacy-preserving multi-centric rapid learning of health care. We also observe some pilot studies applying federated learning and distributed learning to learn patient representations in this review. Li et al. [59] proposed a method that can learn clinical code representations from different EHR databases by adding distributed noise contrastive estimation. Liu et al. [90] proposed a federated learning framework to pre-train on MIMIC-III notes and predict clinical outcomes related to obesity. The adoption of distributed and federated learning shows that there are emerging technologies to manage privacy concerns and facilitate scientific research without directly sharing data.

## 5.C. Limitations

A potential limitation of this review is incomplete retrieval of relevant studies that meet the inclusion criteria. As a wide variety of methods to represent patients are included, it is challenging to conduct an inclusive search strategy with an automatic query by keyword searching. There is, unfortunately, no MeSH term that covers patient representation. We mitigated this issue by working cooperatively with experienced librarians and applying snowballing search strategies from the included publications. It is also challenging to differentiate our included studies with numerous works of predicting one clinical outcome end-to-end with deep learning. To achieve this, we ensured at the full-text review step that all the papers that are included specifically mentioned patient representation or embedding, and that their goal was learning representations at a patient level, not merely a model that maximized performance on a specific disease. In addition, we do not include manuscripts [133–136] that can only be found as preprints on arXiv or bioRxiv. All these limitations would pose a potential threat to selective bias in publication trends.

Another limitation is that our review fails to answer all technical questions. Due to the lack of transparency and reproducibility of reported results, many studies in our review claim state-of-the-art results, but few can be verified by external parties. This might be a barrier for future model development and would slow the pace of improvement. Even though MIMIC-III is widely used as a benchmark dataset, not many generic pipelines for benchmarking machine learning studies are available. Prior works have proposed different components in addition to the deep learning models to improve representation ability and predictive performance. However, it is still unclear which approach works best for representing EHR data. Therefore, one future direction to mitigate this limitation is to more comprehensively investigate different methods and conduct comparative studies using a common set of shared clinical benchmark datasets. Some recent comparative studies such as Ayala Solares et al. [137], Sadati et al. [138], and Min et al. [139] provided some surprising observations in terms of how to choose the best representation learning methods. For instance, Min et al. [139] conducted a case study of applying different machine learning methods to represent Chronic Obstructive Pulmonary Disease (COPD) patient claim data to predict readmission. They have shown some contradictory observations that medical problems are unlike problems in NLP and image processing, merely applying complex deep learning without incorporating medical knowledge does not necessarily result in better performance. Therefore, in the future, we would encourage such studies to be conducted across a variety of patient representation use cases.

## 6. Conclusion

Deep representation learning has led to a wide variety of innovations in the process of modeling EHR patient data. As deep learning models benefit largely from the model capabilities to address the challenges of EHR data, deep patient representation learning is a promising direction to acquire powerful, robust, and precise representations. By adopting advanced learning techniques in addition to the model architecture, patient representation learning attempts to further address issues related to patient data and promote scientific research. We conducted a systematic review of this work and discussed the current research scenarios pertinent to patient representation learning. We believe a growing number of advanced methods will be continuously developed to learn meaningful patient representations, and that these representations will play a greater and greater role in clinical prediction tasks.

## Author Contributions Statement

YS and KR conceived the review. YS, JD, ZL, and KR completed the initial screening. YS and ZL performed data extraction. YS conducted the data analysis and drafted the initial manuscript. XJ, TM, FW, and WJZ contributed to the design of the study and editing of the manuscript. KR provided overall leadership for the study and revised the manuscript. All authors approved the final version of the article.

**Conflicts of Interest**

None declared

**Acknowledgments**

This work was supported by the U.S. National Library of Medicine, National Institutes of Health (NIH) under award R00LM012104, R01LM012973, UL1TR00316701, U01TR002062, R01GM118609, and R01AG06674901, the National Science Foundation under award NSF IIS-1750326, the Cancer Prevention Research Institute of Texas (CPRIT) under awards RP170668 and RR180012, as well as the Patient-Centered Outcomes Research Institute (PCORI) under award ME-2018C1-10963.

**Supplemental Tables**

STable 1.  Proposed Method and Main Contributions for 49 Included Papers

| DL Models | Paper | Proposed Method and Main Contributions | Evaluation Tasks |
|---|---|---|---|
| LSTM | [55] | Compare LSTM and CNN | Intervention Prediction such as invasive ventilation |
| | [69] | BiLSTM-NegTag-Dense-Structured | Disease Prediction including Heart Failure, Stroke, Kidney Failure |
| | [54] | Multiple patient input data with different neural network (CNN for notes and vitals; LSTM for medication; MLP for test order) | Multi-label Disease Prediction and Lab Test Order Prediction |
| | [57] | SHiP: Sequential, Hierarchical and Pretrained LSTM language model | Inpatient Mortality Prediction and Disease Prediction including Diabetes, Heart Failure, and Cancer |
| | [64] | LSTM-attention with a time decay factor | next visit primary diagnosis of 18 major ICD groups |
| | [65] | Heterogeneous Event LSTM with new gate to filter events | Mortality Prediction and Abnormal Lab Test Prediction |
| | [66] | MetaPred: Meta learning-based CNN or LSTM prediction | Disease Prediction of MCI, Alzheimer, Parkinson with 3 digit ICD-9 code |
| | [6] | FHIR-based representations with deep learning architecture including RNN, ATNN, and boosted time-series model | Mortality, Disease (full sets of diagnosis codes from CCS codes), Readmission, Length-of-stay Prediction. |
| CNN | [73] | CNN with multi-task learning | Mortality and Length-of-stay Prediction |
| | [70] | CNN with similarity metric learning | Disease (including Diabetes, COPD, Obesity) Predictions and Patient similarity clustering |
| | [71] | CNN with word2vec feature embedding | Disease (Diabetes and Heart Failure) Predictions |
| | [33] | HCNN: Attributed Graph Representation with Heterogeneous CNN | Co-occurrence of combinations predictions from CHF, Diabetes, CKD, COPD |
| | [14] | RAIM: CNN with attention mechanism of lab and intervention waveforms | 24hr decompensation and Length-of-stay Prediction |
| | [63] | Pretraining of Phenotype-specific text encoder with CNN | Obesity Comorbidities Predictions and Alcohol Misuse Prediction |
| GRU | [76] | Dipole: RNN-based with different attention mechanisms | Multi-label Disease Prediction (Full sets of ICD-9 codes) |
| | [38] | Doctor AI: Gated Recurrent Units with two hidden layers | Forecasting of next visits' time and the codes assigned based on previous visits |
| | [72] | Health-ATM: CBOW-RNN-TargetAttention-TimeConvNet | Disease (Diabetes and Heart Failure) predictions |
| | [77] | Hierarchical attention GRU with adaptive segmentation | Mortality prediction and ICU admission prediction |
| | [78] | Patient2vec: GRU-based with subsequence-level with an attention mechanism | Hospitalization (after 180 days) prediction |
| | [79] | Topic modeling of the global context and RNN of the local context of diagnosis code | Readmission Prediction, Patient Subtype |
| | [80] | RETAIN: Reverse Time with Visit and Variable level Attention RNN | Heart Failure Prediction |
| | [81] | MiME: Joint training of inherent multilevel structure of EHR data with auxiliary tasks | Heart failure prediction and Sequential disease prediction |
| Auto-encoder | [74] | Time-sensitive hybrid Autoencoder with weighted K-means clustering | Patient Subtype of MCI, Parkinson, and Alzheimer Diseases |
| | [75] | Recurrent Neural Network-based Denoising Autoencoder | 10 comorbidity of heart failure predictions and mortality prediction |
| | [11] | Deep Patient: Stacked denoising autoencoder with three hidden layers | Disease (including Diabetes, Heart Failure, Cancer, etc.) Predictions |
| | [82] | SDAE with different feature selections and interpolations | Length-of-stay predictions |
| | [61] | LSTM-based Autoencoder | Mortality prediction at 24 and 48 hours |
| | [83] | Optimized Stacked Denoising Autoencoder | Alcoholism and Length-of-stay prediction |
| | [67] | Time-aware LSTM autoencoder | Diabetes Prediction and Patient Subtype |
| | [84] | SDAE with patient text representations | Mortality, Disease, Procedure, Gender Prediction |
| FFNN | [88] | Time-Aware Attention Feed-forward NN with Bayesian Ensemble | 30-day readmission prediction |
| | [89] | Multi-task Learning with auxiliary tasks | Disease Prediction of Four Types of high-prevalence disease and low-prevalence disease |
| word2vec | [59] | Distributed Noise Contrastive Estimation with the skip-gram embedding of clinical code | Full sets of disease diagnoses |
| | [56] | Joint Skip-gram embedding of structured code and word in the clinical note | Disease diagnoses: acute liver failure, female breast cancer, Schizophrenic disorders |
| | [85] | disease_procedure2vec | Mortality Prediction, Medical charge and Length-of-stay regression |
| | [86] | Med2vec: Skip-gram of structured code | Full sets of diagnoses codes and severity of the clinical risk |
| | [87] | Prediction-guided vector learning of structured code | Medical charge and Length-of-stay predictions |

| Graph NN | [27] | GRAM: Knowledge Directed Acyclic Graph(DAG) with attention model | Full sets of diagnoses codes |
|---|---|---|---|
| | [91] | HeteroMed: heterogeneous information network with node2vec embedding | Disease predictions (two tasks: 1. Exact code, 2. Disease cohort) |
| | [31] | KAME: Knowledge Graph-based attention mechanisms with GRU | Visit-level and code-level disease predictions |
| | [32] | Graph convolutional network of drug-drug interaction and CNN of patient diagnosis and procedure codes with reinforcement learning | Medicine within 24 hours prediction task |
| Deep Averaging Network | [62] | Deep Averaging Network (DAN) with CUI embedding | 16 obesity-related phenotype predictions |
| | [90] | Federated learning of text representation with DAN | 16 obesity-related phenotype predictions |
| Tensor decomposition | [58] | Collective Nonnegative Tensor Factorization with LSTM-based regularization | Mortality prediction and patient similarity |
| | [48] | Tensor Decomposition with predictive task guidance | Hospitalization Prediction and Medical Expense regression |
| Restricted Boltzmann machines | [60] | EMR-driven nonnegative Restricted Boltzmann machines | Suicide risk stratification |
| Transformer | [43] | SAnD: Attend and Diagnosis with multi-headed attention Transformer | Mortality, Decompensation, Length of Stay, Phenotyping. |

STable2. Local Hospital Data Resource

| Resource | Years | Sample size (patient-level) in the study | Patient data type in the study | Paper |
|---|---|---|---|---|
| Shanghai Shuguang Hospital | 12 | 4,682 | Structured | [75] |
| New York University Langone Center | 4 | 300,000 | Unstructured | [55] |
| Danderyd University Hospital | NA | 610 | Structured | [68] |
| Local Medical Claims EHR data | 2 | 9,528 | Structured | [70] |
| Mount Sinai Health System | 34 | 1.2 million | Structured, Unstructured | [11] |
| Medicaid claims | 3 | 147,810 | Structured | [31,76] |
| Sutter Health | 8 | 263,706 | Structured | [27,38,80,81] |
| University of Virginia Health System | 5 | 33,147 | Structured | [33,78] |
| SEER-Medicare Linked Database | 11 | 93,123 | Structured | [64] |
| National University of Singapore Hospital | Three months | 116 | Structured | [82] |
| Oregon Health Science University Hospital | NA | 17,082 | Structured | [66] |
| California State In-hospital Database | 9 | 80,000 | Structured | [85] |
| Hospital Quality Monitoring System in China | 2013-2015 | 40,000 | Structured | [87] |
| University of California, San Francisco | 2012-2016 | 85,522 | Combined | [6] |
| University of Chicago Medicine | 2009-2016 | 108,948 | Combined | [6] |
| Stanford Translational Research Integrated Database Environment (STRIDE) | 7 | 1,221,40 | Structured | [89] |
| Barwon Health, a large regional hospital in Australia | 2009-2012 | 7,578 | Structured | [60] |
| Others (not providing enough description of dataset) | | | | [48,71,72,79,83,86] |